\pdfoutput=1

\documentclass[11pt]{article}

\usepackage[preprint]{acl}

\usepackage{times}
\usepackage{latexsym}

\usepackage[T1]{fontenc}

\usepackage[utf8]{inputenc}

\usepackage{microtype}

\usepackage{inconsolata}

\usepackage{graphicx}

\usepackage{lipsum}
\usepackage{multirow}
\usepackage{booktabs}
\usepackage{graphicx}
\usepackage[linesnumbered,ruled,vlined]{algorithm2e}
\usepackage{float}
\usepackage{amsmath}
\usepackage{pifont}
\usepackage{amssymb}
\usepackage{colortbl}

\usepackage{xspace}
\usepackage{adjustbox}

\newcommand\LN{\linebreak\noindent}
\newcommand{\cmark}{\ding{51}} 

\newcommand{\DOT}{D\texttt{0}T\xspace}

\title{Diverse and Effective Synthetic Data Generation for \\ Adaptable Zero-Shot Dialogue State Tracking}

\author{
  James D. Finch
  \and
  Jinho D. Choi
  \\
  Department of Computer Science
  \\
  Emory University
  \\
  Atlanta, GA, USA
  \\
  \texttt{\{jdfinch, jinho.choi\}@emory.edu}
  \\
}

\begin{document}
\maketitle

\begin{abstract}
We demonstrate substantial performance gains in zero-shot dialogue state tracking (DST) by enhancing training data diversity through synthetic data generation.
Existing DST datasets are severely limited in the number of application domains and slot types they cover due to the high costs of data collection, restricting their adaptability to new domains.
This work addresses this challenge with a novel, fully automatic data generation approach that creates synthetic zero-shot DST datasets.
Distinguished from previous methods, our approach can generate dialogues across a massive range of application domains, complete with silver-standard dialogue state annotations and slot descriptions.
This technique is used to create the \DOT dataset for training zero-shot DST models, encompassing an unprecedented 1,000+ domains. 
Experiments on the MultiWOZ benchmark show that training models on diverse synthetic data improves Joint Goal Accuracy by 6.7\%, achieving results competitive with models 13.5 times larger than ours.
\end{abstract}

\section{Introduction}
\label{sec:dst_introduction}

A critical task for building task-oriented dialogue (TOD) systems is Dialogue State Tracking (DST), which aims to maintain a structured representation of the key task-related information provided throughout a dialogue. Conventionally, the state representation is composed of a set of task-specific slot-value pairs, where slots are information types provided by a predefined slot schema. While DST has been studied in fully supervised \cite{heck_trippy_2020, xie_correctable-dst_2022, won_break_2023} and few-shot settings \cite{lin_leveraging_2021, shin_dialogue_2022, chen_stabilized_2023}, these settings rely on a substantial amount of labeled training examples within the targeted task domain. To this end, zero-shot DST has recently gained attention, as it requires the DST model to adapt to an unseen target domain for which no training examples are available \cite{gupta_show_2022, wang_divide_2023, heck_chatgpt_2023}.

Leveraging slot descriptions to perform cross-task transfer is shown to be effective for zero-shot DST \cite{lin_leveraging_2021, gupta_show_2022, zhao_description-driven_2022, tavares_learning_2023}. In this approach, a model is trained to interpret the slot descriptions to perform DST using gold supervision in several data-rich domains. During inference, the model interprets new slot descriptions to perform DST in unseen target domains without any training data. However, for this approach to succeed, sufficiently diverse training data must be available to enable the model to generalize and handle new slot types. We hypothesize that existing training data for DST is a bottleneck, as the two most popular datasets for DST training, MultiWOZ \cite{budzianowski_multiwoz_2018} and SGD \cite{rastogi_towards_2020}, only cover 7 and 16 domains, respectively.

This work aims to explore the impact of increasing training data diversity on zero-shot DST performance. Since traditional methods of creating diverse DST training data are costly and difficult to scale, we develop a novel, fully automatic data generation approach for zero-shot DST. This approach leverages the capabilities of instruction-tuned large language models (LLMs) to create new task domains from scratch. Synthetic dialogues are generated for each domain, and are automatically annotated for dialogue state, complete with descriptions of labeled slots. This approach is leveraged to generate a synthetic DST dataset of unprecedented diversity, including over $1,000$ task domains. Experiment results demonstrate a substantial performance boost provided by this synthetic data on standard benchmarks.
In summary, our contributions are:

\begin{enumerate}
    \item A novel approach for generating domain-diverse DST data.
    \item A synthetic DST dataset with $1,000+$ domains for training zero-shot models.
    \item Efficient state-of-the-art models that robustly handle diverse domains for zero-shot DST.
\end{enumerate}

\noindent We make all models, code, and data publicly available to support future work.\footnote{\url{https://github.com/anonymous}}


\section{Related Work}
\label{sec:dst_related_work}

\paragraph{Zero-Shot DST} Current state-of-the-art (SoTA) approaches to zero-shot DST use sequence-to-sequence (S2S) modeling to predict appropriate values given a natural language specification of each slot to track \cite{gupta_show_2022, king_diverse_2023}. Such S2S modeling has been effective for adapting to new slot types, since models can leverage descriptions of a new, unseen slot type via in-context learning (ICL) when making predictions. Recently, models using LLMs have achieved state-of-the-art results on this task due to the excellent zero-shot ability of LLMs \cite{hu_-context_2022, king_diverse_2023}. However, the cost of LLM decoding is often too steep for many task-oriented dialogue (TOD) applications. Thus, ongoing work aims to achieve SoTA results with smaller models using cross-task transfer, where the model is trained on an existing set of task domains before being transferred to the unseen target domain \cite{wang_divide_2023, aksu_prompter_2023}.

\paragraph{DST Data Collection} Successful modeling of a low-cost zero-shot DST model that generalizes to unseen domains depends on the quality and diversity of its training data; however, collecting a training resource that covers diverse TOD domains is costly. The most popular dataset, MultiWOZ, was collected using a wizard-of-oz setup using human participants, yet only covers 7 domains \cite{budzianowski_multiwoz_2018}. The Schema Guided Dialogues (SGD) dataset was created in an attempt to increase the diversity of available DST resources using a rule-based data generation approach, where the final dialogue text was paraphrased by crowdworkers to improve naturalness \cite{rastogi_towards_2020}. Even with this more cost-effective collection technique, SGD only covers 16 domains in its training split. Moreover, both datasets suffer from high inter-domain similarity. In the case of MultiWOZ, each domain covers a component of a travel planning application, in which a user talks to an artificial travel agent. As a result, there is a high degree of topical and structural similarity between dialogues, and all domains share a similar focus on scheduling. This results in many overlapping slots between domains to cover scheduling details such as dates, times, and locations. SGD has a more diverse array of domains, yet most are similar to MultiWOZ in that they focus on booking and scheduling. In particular, the Bus, Calendar, Event, Flight, Hotel, RentalCar, Service, and Train domains all share this scheduling focus. As a result of this limited diversity and the cost of additional data collection, it is unknown whether the domain coverage of existing DST resources is a bottleneck for training a zero-shot DST model with robust cross-task transfer.


\paragraph{DST Data Generation} Several previous works explore data augmentation methods for improving the diversity of limited DST data. Nearly all of these approaches target the few-shot setting, where a limited number of labeled examples are used as a seed set to be augmented with additional, synthetic examples. This can be done using simple approaches to improve the lexical \cite{quan_effective_2019, yin_dialog_2020} or semantic \cite{summerville_how_2020, lai_controllable_2022} diversity of training examples, or by synthesizing entire dialogues \cite{campagna_zero-shot_2020, aksu_velocidapter_2021, aksu_n-shot_2022, mehri_lad_2022, mohapatra_simulated_2021, kim_neuralwoz_2021, wan_unified_2022} to create additional training resources.
These previous works in DST data generation demonstrate that automatic methods for data augmentation and generation can help address the limitations of existing training resources and improve transfer to data-poor domains. Additional detail regarding related work in DST data generation is provided in Appendix \ref{appx:related_work_for_dst_data_generation}. 

Our DST data generation approach is distinct from all previous methods because it generates entirely new task domains, in addition to new dialogues with silver annotations. Furthermore, our approach is fully automatic, requiring no few-shot data or manual creation of domain-specific resources, making it ideal for scaling up the diversity of training resources for zero-shot DST.

\section{DST Data Generation}
\label{sec:generating_dst_data}

This section presents our fully automatic data generation approach to support training DST models capable of zero-shot domain transfer.
Our goal is to create a set of dialogue data covering many diverse task domains, with silver dialogue state labels and natural language slot descriptions.
Given the exceptional zero-shot performance of instruction-tuned large language models (LLMs) on a wide variety of tasks \cite{brown_language_2020, kojima_large_2022, heck_chatgpt_2023}, our approach explores using instruction-tuned LLMs for data generation. We use GPT\footnote{gpt-3.5-turbo-0301 is used for all stages of the approach, except for QA Pair Generation in which gpt-4-0314 is used.} in all of our presented experiments, although any LLM can be used for our approach in principle.

\begin{figure}[htp!]
    \centering
    \includegraphics[width=\columnwidth]{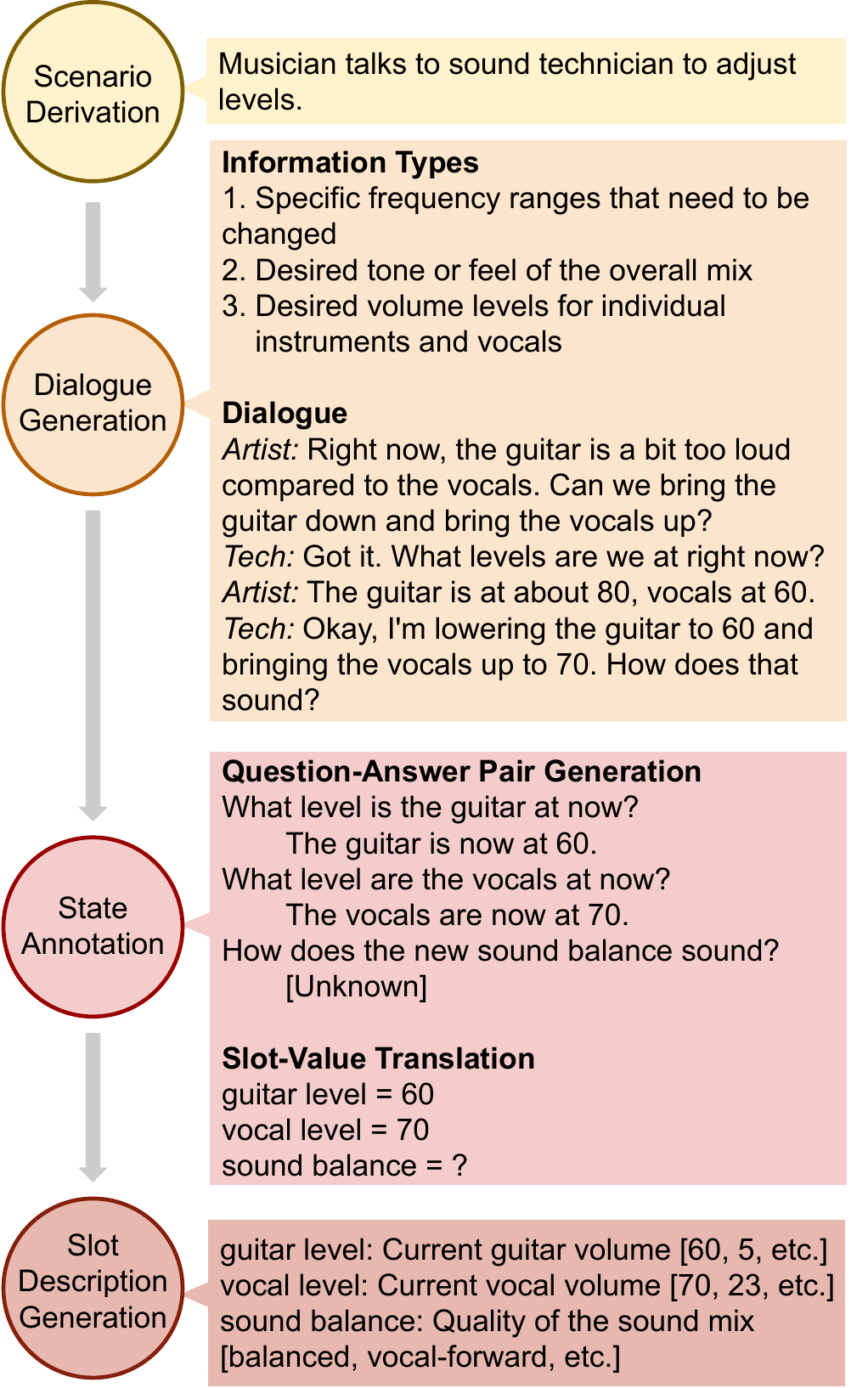}
    \caption{The four-stage DST data generation pipeline.}
    \label{fig:gendis}
    \vspace{-2ex}
\end{figure}

The approach consists of four stages, which are summarized in Figure \ref{fig:gendis}.
First, domains are derived through an iterative process of generating and refining dialogue scenario descriptions (\textsection\ref{ssec:scenario-derivation}).
Next, a dialogue is crafted based on the scenario description and a generated unstructured information list corresponding to the scenario (\textsection\ref{ssec:dialogue-generation}).
Third, each turn in each dialogue is automatically annotated with silver dialogue state labels (\textsection\ref{ssec:automatic_state_annotation}).
Finally, a slot description is composed for each silver slot-value pair annotation (\textsection\ref{ssec:slot_description_generation}).
All prompts included in the approach are provided in Appendix~\ref{appx:prompts}.

\subsection{Scenario Derivation}
\label{ssec:scenario-derivation}

\noindent Algorithm~\ref{alg:scenario-gen} shows our scenario derivation method.
GPT is iteratively prompted to create a mini-set of $k$ dialogue scenario descriptions (\texttt{L3}).
Each mini-set is combined with the scenarios obtained from previous iterations, where each scenario description is encoded into an embedding by SentenceBERT\footnote{SentenceBert model: \texttt{all-MiniLM-L6-v2}} \cite{reimers_sentence-bert_2019} and the resulting embeddings are clustered through a community detection algorithm (\texttt{L4}).\footnote{\href{https://www.sbert.net/docs/package_reference/util.html}{https://www.sbert.net/docs/package\_reference/util.html}} A deduplicated set of scenario descriptions is created by selecting \textit{one} embedding from every cluster, which is mapped back to its corresponding scenario description (\texttt{L5}).
This iteration continues until the set reaches the requested size (\texttt{L2}). In our case, $k = 100, n = 1000$.
Appx. \ref{appx:domains} gives a sample of the generated scenarios.

\begin{algorithm}
    \caption{Scenario Derivation}
    \label{alg:scenario-gen}
    \fontsize{10.3}{5.0}\selectfont
    \SetKwInOut{Input}{Input}
    \SetKwInOut{Output}{output}
    \Input{$k$: mini-set size, $n$: final set size.}
    \Output{$S$: the final set containing $n$ scenarios.}
    $S \leftarrow \emptyset$ \\
    \While{|$S$| $<$ $n$}{
        $S' \leftarrow \textbf{gpt\_generated\_scenarios}(k)$ \\
        $\mathbb{E} \leftarrow \textbf{cluster}(\textbf{embed}(S \cup S'))$ \\
        $S \leftarrow \{\forall_{C \in \mathbb{E}}.\: \textbf{map}({\textbf{one}}(c)) : c \in C\}$ \\
    }
    \Return $S$
\end{algorithm}


\subsection{Dialogue Generation}
\label{ssec:dialogue-generation}

In a pilot analysis, generating dialogues directly from scenario descriptions (\textsection\ref{ssec:scenario-derivation}) using GPT resulted in generic contents that lack sufficient details for effective DST model training.
To address this issue, we generate dialogues from scenario descriptions in two steps. First, GPT is asked to generate a comprehensive list of information types based on the provided scenario, which serves as a de-facto ontology for representing the properties of the scenario.
Second, given a scenario and its associated information types, GPT is then asked to generate a dialogue. The prompt encourages GPT to provide detailed responses and make up values for the information types in order to encourage generating concrete values to serve as targets for DST.

\begin{figure*}[htb]
    \centering
    \includegraphics[width=\textwidth]{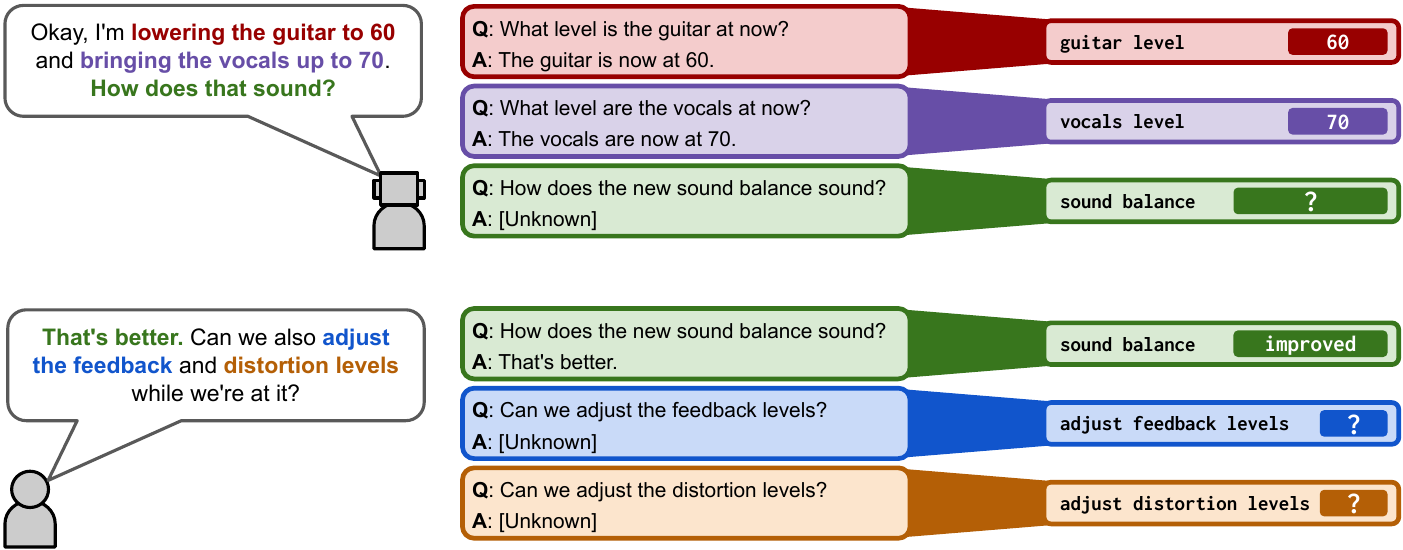}
    \caption{Example turn outputs from the automatic state annotation component of the DST data generation pipeline.}
    \label{fig:dialqasv}
    \vspace{-0.5em}
\end{figure*}

\subsection{State Annotation}
\label{ssec:automatic_state_annotation}

Each turn in the generated dialogues is automatically annotated with a dialogue state update using two components: \textit{Question-Answer (QA) Pair Generation} to deduce the key information in each turn and \textit{Slot-Value Translation} to transform those QA pairs into slot names and values. 
Figure~\ref{fig:dialqasv} illustrates the automatic state annotation approach.

\paragraph{Question-Answer Pair (QA) Generation}

To generate a state update $U_t$ given a dialogue history $D_{1..t}$, we use a prompt $P^{QA}_t$ containing the last two turns $D_{t-1,t}$, and instruct GPT to break down all the information in turn $t$ as a set of QA pairs.
Only the last two turns are included to reduce irrelevant information from previous turns that could misguide the state update for the current turn $t$.
To further mitigate this issue, every turn is prepended with a speaker tag, allowing GPT to soley focus on turn $t$ by referring to the corresponding speaker.
A set of QA pairs $QA_t = \{(q^t_1, a^t_1), \ldots, (q^t_k, a^t_k)\}$ is generated by this method, where each question~$q^t_i$ represents an information type either shared or requested during the turn and its answer $a^t_i$ summarizes the information value.

State updates are produced to monitor the change in values of slots throughout the dialogue, enabling us to track whether information requests from one speaker are satisfied through information shared by the other speaker. 
To implement this, $P^{QA}_t$ explicitly designates the answer \textit{Unknown} for use in any QA pair, where the question represents an information request made by the current speaker.
Therefore, for each turn, a set of unanswered questions for the prompt $P^{QA}_t$ can be identified as follows:
$$R_t = \{\forall_i.\: q^t_i : 0 < i \leq k \land a^t_i = \textit{Unknown}\}$$

\noindent A second prompt $P^{A}$ is used to answer each question in $R_t$ using two turns $D_{t, t+1}$, which produces a set of QA pairs $QA'_{t+1}$ comprising slots from turn $t$ filled with values in turn $t+1$.
Included in $P^{A}$ is an instruction to use \textit{Unknown} for questions whose answers are not present in turn $t+1$.
Such unanswered questions are removed from $QA'_{t+1}$, leaving only\LN QA pairs with information requested in turn $t$ and shared in turn $t+1$. 
$QA'_{t+1}$ are then appended to the next prompt $P^{QA}_{t+1}$ to generate a new set $QA_{t+1}$ for turn $t+1$.
Including $QA'_{t+1}$ in $P^{QA}_{t+1}$ guides GPT to generate only new QA pairs that have not already been covered by $QA'_{t+1}$.

\paragraph{Slot-Value Translation}

After summarizing key dialogue information as QA pairs, every QA pair in $QA_t$ is translated to a slot-value pair.
GPT tends to generate overly detailed slot names when answers are provided along with questions. 
Hence, slot names and values are derived using separate prompts. 
First, a prompt $P^{S}$ is used to translate all questions in $QA_t$ into corresponding slot names.
No context from the dialogue is provided, nor do we include any answers from $QA_t$ in $P^{S}$.
The result is a set of slot names $N_t = \{s^t_1, \ldots, s^t_{|QA_t|}\}$ representing information types mentioned in turn $t$.

Finally, a prompt $P^{V}$, comprising questions and answers in $QA_t$ as well as the slot names in $N_t$, is used to translate each answer into a value for the corresponding slot name.
In addition, $P^{V}$ highlights that a value can be a concise phrase, number, span, category, score, boolean, list, or other form, aiding the model in generating values suitable for the respective slot names, rather than always using natural language phrases as values.
QA pairs with the \textit{Unknown} answer are excluded from $P^{V}$, as they are translated into a special token $?$ to represent a requested slot. 
Pairing each generated value with its corresponding slot name results in the dialogue state update $U_t = \{(s^t_1, v^t_1), \ldots, (s^t_{|QA_t|}, v^t_{|QA_t|})\}$.

\subsection{Slot Description Generation}
\label{ssec:slot_description_generation}

For each state update $U_t$ produced by automatic annotation (\textsection\ref{ssec:automatic_state_annotation}), GPT is instructed to generate a specification of each slot in $U_t$ using a single prompt.  The prompt includes each slot value pair $(s^t_i, v^t_i)$ in $U_t$ as well as each question $q^t_i$ corresponding to each slot. GPT is asked to generate a description for each slot as a short natural language phrase $d^t_i$, in addition to a few comma-separated example values $e^t_i$ that could fill the slot.

\section{New Dataset for Zero-Shot Tracking}
\label{sec:dst_dot_dataset}

Using our DST data generation approach (\textsection\ref{sec:generating_dst_data}), we create a Diverse \texttt{0}-shot Tracking dataset: \DOT. Since we aim to measure the impact of increasing the diversity of DST training resources, we generate \DOT to include unprecedented $1,000+$ domains and $5$ dialogues per domain. Applying automatic state annotation (\textsection\ref{ssec:automatic_state_annotation}) to the generated dialogues yields $324,973$ slot-value pairs in state updates. Since compiling each dialogue state $S_t = update(S_{t-1}, U_t)$ produces an excessive $\approx 6.5$ million total slot-value pairs for DST training, slot-value pairs are downsampled using a method that maintains slot type diversity. We randomly sample exactly $1$ example for each of the original $324,973$ slot-value updates from the set of final slot-values where that slot is filled (non-empty), resulting in $n = 324,973$ filled slot-value examples. To include examples of empty slots, we randomly sample $m$ empty slot-value pairs from the final compiled states, where $m = 0.5 * n = 162,487$. Table~\ref{tab:statistics} presents the final statistics of the dataset, and Table \ref{tab:data_comparison} presents a comparison to existing data. 

\begin{table}[htbp!]
  \centering\small
  \begin{tabular}{lr|lr}
      \toprule
      \textbf{Metric} & \textbf{Value} & \textbf{Metric}  & \textbf{Value} \\
      \midrule
      Scenarios       &          1,003 & Unique Slots     &        173,572 \\
      Dialogues       &          5,015 & Unique Slots$_S$ &          244.6 \\
      Turns           &        100,471 & Unique Slots$_D$ &           64.9 \\
      Turns$_D$       &           20.0 & Unique Slots$_T$ &            3.3 \\
      Tokens          &      2,061,332 & Turns w/o SV     &          1,583 \\
      Tokens$_T$      &           20.5 & Tokens$_{SN}$    &            2.4 \\
      Slot-Values     &        487,460 & Tokens$_{SV}$    &            2.0 \\
      \bottomrule
  \end{tabular}
  \caption{The statistics of the \DOT dataset with dialogue state update labels created using our fully automatic generation pipeline (\textsection\ref{sec:generating_dst_data}). SN/SV: slot names/values respectively, *$_{D/T/S/SN/SV}$: * per dialogue/turn/scenario/SN/SV, respectively.}
  \label{tab:statistics}
  \vspace{-1ex}
\end{table}

\noindent We validate the quality of the dataset by recruiting 3 human evaluators to annotate $60$ randomly sampled turns, judging (1) whether each slot-value correctly represents information in the corresponding turn and (2) whether each state update $U_t$ is missing any important information in the turn. $82\%$ of slot-value pairs were judged correct and $7\%$ of state updates were missing important information. 

\begin{table}[ht]
    \centering\small
    \resizebox{\columnwidth}{!}{%
    \begin{tabular}{c|ccccc}
        \toprule
        Dataset & Dom. & Dial. & Turns & SV & US \\
        \midrule
        MWOZ & 7 & 8,438 & 113,556 & 4,510 & 24 \\
        SGD & 16 & 16,142 & 329,964 & 14,139 & 214 \\
        \DOT & 1,003 & 5,015 & 100,471 & 487,460 & 173,572 \\
        \bottomrule
    \end{tabular}
    }
    \caption{Comparison of \DOT to the train splits of MultiWOZ 2.1/2.4 (MWOZ) and SGD, compared on number of domains (Dom.), dialogues (Dial.), turns, slot-values (SV), and unique slot names (US).}
    \label{tab:data_comparison}
    \vspace{-2ex}
\end{table}

\section{Experiment Setup}
\label{sec:dst-experiments}

\paragraph{Evaluation Data} 

Our experiments on zero-shot DST use the standard MultiWOZ benchmark \cite{budzianowski_multiwoz_2018}. This evaluation was designed using a leave-one-out setup in which a zero-shot DST model is tested on each of five domains (Attraction, Hotel, Restaurant, Taxi, Train) after being trained on the other four, to test zero-shot transfer to new domains. Joint Goal Accuracy (JGA) is the evaluation metric, measuring the proportion of turns for which the entire dialogue state is correctly inferred. The MultiWOZ 2.4 \cite{ye_multiwoz_2022} variant is used as the main evaluation dataset since it contains corrected gold labels in the validation and test splits. We additionally include an evaluation on the uncorrected MultiWOZ 2.1 variant \cite{eric_multiwoz_2020} to facilitate further comparison to previous work. 

Since MultiWOZ does not contain slot descriptions, a single-sentence description is written for each MultiWOZ slot to provide slot definitions. Descriptions are authored based on \citet{lin_leveraging_2021} but with improvements in detail and grammar. Additionally, descriptions are augmented with $4$ value examples for each slot. No prompt engineering or validation experiments are performed when creating slot descriptions and value examples, to reflect the performance of the model in real-world settings without requiring extensive development effort.

\paragraph{Models}

The impact of domain-diverse training data on zero-shot DST is evaluated by comparing models that leverage the domain-diverse \DOT dataset as a training resource against baselines trained only on the standard training splits of benchmark data. Models leveraging \DOT (\texttt{+\DOT}) are trained in two sequential training stages. Models are first trained on \DOT to acquire domain-general state tracking ability, and then refined in a second training stage using the standard training split of benchmark data. 

\begin{figure}[ht]
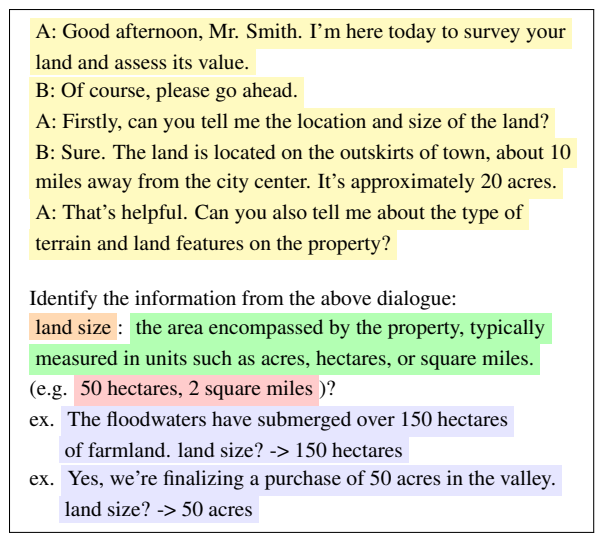

    \centering
    \adjustbox{fbox, max width=\columnwidth}{%
    \resizebox{\columnwidth}{!}{%
    \begin{tabular}{p{1.25\columnwidth}}
        \colorbox{yellow!30}{A: Good afternoon, Mr. Smith. I'm here today to survey your} \\
        \colorbox{yellow!30}{land and assess its value.} \\
        \colorbox{yellow!30}{B: Of course, please go ahead.} \\
        \colorbox{yellow!30}{A: Firstly, can you tell me the location and size of the land?} \\
        \colorbox{yellow!30}{B: Sure. The land is located on the outskirts of town, about 10} \\
        \colorbox{yellow!30}{miles away from the city center. It's approximately 20 acres.} \\
        \colorbox{yellow!30}{A: That's helpful. Can you also tell me about the type of} \\
        \colorbox{yellow!30}{terrain and land features on the property?} \\
        \\
        Identify the information from the above dialogue: \\
        \colorbox{orange!30}{land size}: \colorbox{green!30}{the area encompassed by the property, typically} \\
        \colorbox{green!30}{measured in units such as acres, hectares, or square miles.} \\
        (e.g. \colorbox{red!20}{50 hectares, 2 square miles})? \\
        ex. \colorbox{blue!10}{The floodwaters have submerged over 150 hectares} \\ 
        \;\;\;\; \colorbox{blue!10}{of farmland. land size? -> 150 hectares} \\
        ex. \colorbox{blue!10}{Yes, we're finalizing a purchase of 50 acres in the valley.} \\ 
        \;\;\;\; \colorbox{blue!10}{land size? -> 50 acres} 
    \end{tabular}
    }}
    \caption{An example of an input token sequence from the \DOT dataset used for training. [\texttt{YELLOW}]: dialogue context $D_{1..t}$\;\; [\texttt{PEACH}]: slot $s^t_i$\;\; [\texttt{GREEN}]: slot description $d^t_i$\;\; [\texttt{RED}]: value examples $e^t_i$\;\; [\texttt{BLUE}]: In-context demonstrations (\texttt{+ICL} only)}
    \label{fig:dst-sequence-format}
    \vspace{-1ex}
\end{figure}

Two base models, T5 1.1 \cite{raffel_exploring_2020} and Llama2-Chat \cite{ouyang_training_2022}, are used in our experiments. We use the 11B and 13B variants of the T5 and Llama2 models, respectively; however, for greater efficiency and robustness for two-stage model training, we additionally leverage the QLoRA \cite{dettmers_qlora_2023} quantization and training method. 
Models are trained using the sequence-to-sequence format shown in Figure \ref{fig:dst-sequence-format} which follows the "independent" formulation from \citet{gupta_show_2022}.
Appendix \ref{appx:implementation-details} provides implementation details such as model hyperparameters.

\noindent Additionally, since recent work in zero-shot DST has shown performance improvements from including demonstrations in slot descriptions using in-context learning \cite{gupta_show_2022, hu_-context_2022, king_diverse_2023}, we also experiment with this approach using the Llama2 base model, to observe the interaction between domain-diverse training and in-context demonstration. 
Models leveraging in-context demonstrations (\texttt{+ICL}) are trained and tested with slot descriptions that include up to $k=3$ in-context demonstrations, where $k$ is a per-domain hyperparameter selected by validation performance.

For MultiWOZ, demonstrations are collected for each slot by manually constructing $3$ single-turn examples of the slot being updated with an appropriate value.
For \DOT, we collect in-context demonstrations using a fully automatic method in order to preserve the fully-automatic nature of the data generation approach. This is done by augmenting slot descriptions in the \DOT dataset by sampling slot-value labels that share similar semantics to the target slot. 
Similar slot-value examples are found for demonstration sampling by encoding every silver slot-value update label in \DOT as the token sequence "$s\text{: }v$" using SBERT \cite{reimers_sentence-bert_2019} and then clustering the encoded slot-values using HDBSCAN \cite{mcinnes_hdbscan_2017}.
Then, for each training example of slot name, value, and slot description $(s,v,d)$, up to $3$ demonstrations are randomly sampled from other training examples that appear in the same cluster and the same domain, but different dialogues.
The description $d$ is augmented by appending each sampled demonstration value with the text of the dialogue turn in which it appears, using the format exemplified in Figure \ref{fig:dst-sequence-format}. 
\section{Results}
\label{sec:dst-results}

\begin{table*}[!ht]
    \small
    \centering
    \resizebox{\textwidth}{!}{
    \begin{tabular}{lllllllll}
    \toprule
        \textbf{data} & \textbf{model} & \textbf{params} & \textbf{avg.} & \textbf{attr.} & \textbf{hotel} & \textbf{rest.} & \textbf{taxi} & \textbf{train} \\ \toprule
        \multirow{8}{*}{MWOZ 2.4} & IC-DST \cite{hu_-context_2022} & 175B & 58.7 & 62.1 & 53.2 & 54.9 & 71.9 & 51.4 \\ 
        ~ & ParsingDST \cite{wu-etal-2023-semantic} & 175B & 64.7 & 65.6 & 46.8 & 67.7 & 80.6 & 62.6 \\
        ~ & RefPyDST \cite{king_diverse_2023} & 175B & 68.8 & 74.5 & 56.6 & 68.2 & 68.5 & 76.1 \\
        \cmidrule(lr){2-9}
        ~ & T5-QLoRA & 11B & 47.1 & 63.9 & 24.1 & 65.5 & 29.4 & 52.9 \\ 
        ~ & \cellcolor{yellow!20}\;\;\;\;+\DOT & \cellcolor{yellow!20}11B & \cellcolor{yellow!20}55.7 (+8.6) & \cellcolor{yellow!20}68.1 & \cellcolor{yellow!20}32.0 & \cellcolor{yellow!20}72.3 & \cellcolor{yellow!20}50.6 & \cellcolor{yellow!20}55.8 \\ 
        ~ & Llama2-QLoRA & 13B & 59.2 & 62.2 & 44.9 & 69.8 & 49.1 & 70.2 \\
        ~ & \;\;\;\;+ICL & 13B & 62.0 (+2.8) & 74.7 & 44.9 & 69.8 & 49.1 & 71.3 \\
        ~ & \cellcolor{yellow!20}\;\;\;\;+\DOT & \cellcolor{yellow!20}13B & \cellcolor{yellow!20}65.9 (+6.7) & \cellcolor{yellow!20}74.4 & \cellcolor{yellow!20}56.4 & \cellcolor{yellow!20}76.0 & \cellcolor{yellow!20}54.7 & \cellcolor{yellow!20}68.3 \\ 
        ~ & \cellcolor{yellow!20}\;\;\;\;+\DOT+ICL & \cellcolor{yellow!20}13B & \cellcolor{yellow!20}68.6 (+9.4) & \cellcolor{yellow!20}76.8 & \cellcolor{yellow!20}56.4 & \cellcolor{yellow!20}78.8 & \cellcolor{yellow!20}54.7 & \cellcolor{yellow!20}76.1 \\ 
        \midrule
        \multirow{11}{*}{MWOZ 2.1} & D3ST \cite{zhao_description-driven_2022} & 11B & 46.7 & 56.4 & 21.8 & 38.2 & 78.4 & 38.7 \\ 
        ~ & ChatGPT \cite{heck_chatgpt_2023} & 175B & 56.4 & 52.7 & 42.0 & 55.8 & 70.9 & 60.8 \\
        ~ & IC-DST \cite{hu_-context_2022}  & 175B & 57.0 & 60.0 & 46.7 & 57.3 & 71.4 & 49.4 \\
        ~ & ParsingDST \cite{wu-etal-2023-semantic} & 175B & 63.4 & 65.0 & 46.8 & 67.0 & 80.3 & 62.8 \\
        ~ & RefPyDST \cite{king_diverse_2023} & 175B & 64.7 & 70.9 & 51.2 & 65.6 & 67.1 & 69.2 \\ 
        ~ & SDT \cite{gupta_show_2022} & 11B & 65.9 & 74.4 & 33.9 & 72.0 & 86.4 & 62.9 \\
        \cmidrule(lr){2-9}
        ~ & T5-QLoRA & 11B & 42.6 & 55.7 & 20.8 & 60.7 & 27.2 & 48.7 \\ 
        ~ & \cellcolor{yellow!20}\;\;\;\;+\DOT & \cellcolor{yellow!20}11B & \cellcolor{yellow!20}49.9 (+7.3) & \cellcolor{yellow!20}61.1 & \cellcolor{yellow!20}27.6 & \cellcolor{yellow!20}64.3 & \cellcolor{yellow!20}46.9 & \cellcolor{yellow!20}49.7 \\ 
        ~ & Llama2-QLoRA & 13B & 51.8 & 55.4 & 38.8 & 59.0 & 44.8 & 61.2 \\ 
        ~ & \;\;\;\;+ICL & 13B & 54.0 (+2.2) & 63.8 & 38.8 & 59.0 & 44.8 & 63.5 \\
        ~ & \cellcolor{yellow!20}\;\;\;\;+\DOT & \cellcolor{yellow!20}13B & \cellcolor{yellow!20}56.2 (+4.4) & \cellcolor{yellow!20}63.1 & \cellcolor{yellow!20}43.8 & \cellcolor{yellow!20}64.7 & \cellcolor{yellow!20}48.8 & \cellcolor{yellow!20}60.8 \\ 
        ~ & \cellcolor{yellow!20}\;\;\;\;+\DOT+ICL & \cellcolor{yellow!20}13B & \cellcolor{yellow!20}58.5 (+6.7) & \cellcolor{yellow!20}66.6 & \cellcolor{yellow!20}43.8 & \cellcolor{yellow!20}67.2 & \cellcolor{yellow!20}48.8 & \cellcolor{yellow!20}66.5 \\ 
        \bottomrule
    \end{tabular}
    }
    \caption{Zero-shot DST results on MultiWOZ (JGA). Parentheses indicate the difference in performance compared to the baseline within base model groups. +\texttt{\DOT} indicates training on \DOT in an initial stage of training. +\texttt{ICL} indicates use of in-context demonstrations.}
    \label{tab:dst_mwoz_results}
    \vspace{-2ex}
\end{table*}

\paragraph{Impact of Domain-Diverse Training}

Table \ref{tab:dst_mwoz_results} presents the results of the zero-shot DST evaluation. Training on the domain-diverse synthetic dataset \DOT results in substantial performance gains across all models. On MultiWOZ 2.4, T5 and Llama2 gain +$8.6$ and +$6.7$ average JGA respectively. Gains on MultiWOZ 2.1 are more moderate at +$7.3$ for T5 and +$4.4$ for Llama2, which is expected as noisy gold labels make improvements less observable. 

Interestingly, our models benefit from the gold label corrections of MultiWOZ 2.4 more than previous approaches. Llama2 \texttt{+\DOT} \texttt{+ICL} benefits the most of any model from the MultiWOZ 2.4 corrections, indicating that it is punished for a substantial amount of correct predictions on MultiWOZ 2.1. 

Llama2 demonstrated far better performance than T5 for both baseline and \texttt{+\DOT} settings.
With the improvements from \DOT training, our Llama2 models achieve performance that is competitive with approaches based on language models of much larger ($\approx175$ billion) parameter counts such as ChatGPT3.5 \cite{heck_chatgpt_2023, wu-etal-2023-semantic} and OpenAI Codex \cite{hu_-context_2022, king_diverse_2023}, and our best Llama2 \texttt{+\DOT} \texttt{+ICL} model is within $0.2$\% of the current SoTA.

\paragraph{Impact of In-Context Demonstrations}

Adding in-context demonstrations to slot descriptions results in a consistent $2$-$3$\% performance gain for both \texttt{+\DOT} and baseline Llama2 models. This is consistent with previous work that tests the impact of in-context demonstrations \cite{gupta_show_2022}. Encouragingly, the performance benefits of \texttt{+ICL} and \texttt{+\DOT} appear to stack, yielding a combined improvement of +$9.4$ average JGA on MultiWOZ 2.4.
\vspace{-2ex}

\paragraph{Comparison of Domain-Diverse Data}
\label{ssec:impact_of_similar_domains}

To further verify the effectiveness of \DOT as a domain-diverse training resource, we compare against the most domain-diverse existing dataset, Schema-Guided Dialogues (SGD) \cite{rastogi_towards_2020}. We train a Llama2 model using the entire SGD training split as a first training stage to replace \DOT training, before fine-tuning on MultiWOZ in the second stage to make a direct comparison. As shown in Table \ref{tab:filtered-performance}, the model leveraging \DOT training outperforms a model that utilizes SGD instead. This demonstrates the power of the massively increased domain diversity covered by \DOT, despite it being a synthetic dataset created with no human intervention. This result also validates the effectiveness of our automatic generation pipeline since it can yield useful training resources while only incurring a small fraction of the time and cost compared to traditional data collection methods.

\begin{table}[!ht]
    \small
    \centering
    \resizebox{\columnwidth}{!}{
    \begin{tabular}{ll|l|lllll}
    \toprule
    \textbf{TD} & \textbf{F} & \textbf{avg.} & \textbf{attr.} & \textbf{hotel} & \textbf{rest.} & \textbf{taxi} & \textbf{train} \\ 
    \midrule
    SGD &  & 65.1 & 76.0 & 51.6 & 76.8 & 53.5 & 68.0 \\
    SGD & \cmark & 61.8 & 75.6 & 45.1 & 77.0 & 46.8 & 64.5 \\
    \DOT &  & 65.9 & 74.4 & 56.4 & 76.0 & 54.7 & 68.3 \\
    \DOT & \cmark & 66.3 & 78.8 & 53.9 & 75.0 & 53.0 & 71.1 \\
    \bottomrule
    \end{tabular}
    }
    \caption{Zero-shot DST results on MultiWOZ 2.4 (JGA), comparing the efficacy of \DOT versus SGD as a domain-diverse resource for stage one training. Llama2 is used as a base model with QLoRA training. TD: Stage one training dataset. F: Checked if domains similar to MultiWOZ are filtered out before training.}
    \label{tab:filtered-performance}
    \vspace{-2.5ex}
\end{table}

One limitation of evaluating SGD as a domain-diverse training resource on the MultiWOZ benchmark is that SGD contains an approximate superset of the domains in MultiWOZ. Consequently, the ability of SGD to train a domain-generalizable DST model is not tested. To address this, we simulate the effectiveness of SGD to improve zero-shot performance for new domains by filtering out all training examples that belong to a domain analogous to those seen in MultiWOZ. Specifically, we filter out the Travel, Hotel, Restaurant, RideShare, and Trains domains and train another baseline model using this filtered datatset. As shown in Table \ref{tab:filtered-performance}, zero-shot performance is impacted by -$3.3$ average JGA as a result of this filtering. Although \DOT can be trivially extended to new domains using our automatic data generation pipeline, we similarly test its capability for training models that generalize to new domains by training a model using a filtered version of \DOT. Filtering is performed by manually reviewing all $1,003$ domains and excluding any that include attractions, hotels, restaurants, taxis, trains, or general travel planning as a primary theme. Model performance remains virtually identical (+$0.4$) regardless of whether \DOT domains are filtered based on similarity to MultiWOZ domains, which is evidence that the benefits of training on \DOT generalize to unseen domains.

\paragraph{Impact of Trainable Parameter Size}

We investigate the interaction between the parameter efficient training technique QLoRA and domain-diverse training by evaluating a variant of our T5 model with full finetuning and without quantization (i.e. without QLoRA). Additionally, a 3 billion T5 base model is compared to evaluate the impact of model size. Results are presented in Table \ref{tab:training_configuration_results}. 
Consistent with previous work, we find that increasing model size yields substantial performance improvements on zero-shot DST. Whereas the T5-3B benefits from training on \DOT, we observe a slight performance loss when training T5-11B, likely due to catastrophic forgetting when training on noisy \DOT labels. Although QLoRA appears to moderately harm performance when training the T5-11B baseline, the T5-11B-QLoRA model actually achieves the best overall performance when first trained on \DOT, likely due to the ability of QLoRA to protect against catastrophic forgetting.  

\begin{table}[!ht]
    \small
    \centering
    \resizebox{\columnwidth}{!}{
    \begin{tabular}{l|l|lllll}
    \toprule
         \textbf{model} & \textbf{avg.} & \textbf{attr.} & \textbf{hotel} & \textbf{rest.} & \textbf{taxi} & \textbf{train} \\ \midrule
         3B & 49.2 & 63.2 & 26.0 & 71.7 & 29.8 & 55.8 \\ 
         \cellcolor{yellow!20}\;\;\;\;$+$\DOT  & \cellcolor{yellow!20}51.5 & \cellcolor{yellow!20}69.1 & \cellcolor{yellow!20}29.9 & \cellcolor{yellow!20}73.2 & \cellcolor{yellow!20}29.2 & \cellcolor{yellow!20}56.2 \\
         11B & 53.8 & 65.0 & 27.6 & 71.0 & 37.5 & 68.2 \\ 
         \cellcolor{yellow!20}\;\;\;\;+\DOT & \cellcolor{yellow!20}52.4 & \cellcolor{yellow!20}70.3 & \cellcolor{yellow!20}29.1 & \cellcolor{yellow!20}66.8 & \cellcolor{yellow!20}36.1 & \cellcolor{yellow!20}59.9 \\
         11B-QLoRA & 47.1 & 63.9 & 24.1 & 65.5 & 29.4 & 52.9 \\ 
         \cellcolor{yellow!20}\;\;\;\;+\DOT & \cellcolor{yellow!20}55.7 & \cellcolor{yellow!20}68.1 & \cellcolor{yellow!20}32.0 & \cellcolor{yellow!20}72.3 & \cellcolor{yellow!20}50.6 & \cellcolor{yellow!20}55.8 \\ 
        \bottomrule
    \end{tabular}
    }
    \caption{Zero-shot DST results on MultiWOZ 2.4 (JGA), comparing 3B, 11B, and 11B-QLoRA variants of the T5 base model. +\texttt{\DOT} indicates training on \DOT in an initial stage of training.}
    \label{tab:training_configuration_results}
    \vspace{-3ex}
\end{table}

\paragraph{Analysis of Training Stages}
\label{ssec:out_of_distribution_generalization}

The efficacy of \DOT as a training dataset for zero-shot DST is further investigated by comparing the performance of the Llama2 model at the conclusion of each stage of training. Table \ref{tab:model-performance-2} presents results on the MultiWOZ 2.4 benchmark for the stage one model trained only on \DOT versus the stage two model additionally trained on MultiWOZ. As expected, the second stage of training is revealed to be crucial as the stage one model achieves only $23.6\%$ average JGA. This reflects the effect of training on noisy dialogue state labels produced by automatic generation, which humans judged to have a slot-value pair correctness rate of $82\%$\footnote{Note that JGA is a more punishing metric than the percent of correct slot-values}. Taken together with the results in Table \ref{tab:filtered-performance}, this result suggests that the benefit provided by \DOT is due to its diversity rather than its overall quality compared to existing data. Further refinements to the automatic data generation pipeline presented in Section \ref{sec:generating_dst_data} to generate more accurate state labels may yield additional performance gains. An error analysis of stage one and stage two models is provided in Appendix \ref{appx:error-analysis}. 

\begin{table}[!ht]
    \small
    \centering
    \resizebox{\columnwidth}{!}{
    \begin{tabular}{c|l|rrrrrr}
    \toprule
     \textbf{Stage} & \textbf{avg.} & \textbf{attr.} & \textbf{hotel} & \textbf{rest.} & \textbf{taxi} & \textbf{train} \\ 
    \midrule
    1 & 23.6 & 26.7 & 11.4 & 39.7 & 13.9 & 26.9 \\
    2 & 65.9 & 74.4 & 56.4 & 76.0 & 54.7 & 68.3 \\
    \bottomrule
    \end{tabular}
    }
    \caption{Zero-shot DST results on MultiWOZ 2.4 (JGA), comparing Llama2 with QLoRA after training only on \DOT (Stage 1) versus after additionally training on MultiWOZ (Stage 2).}
    \label{tab:model-performance-2}
    \vspace{-2ex}
\end{table}

\section{Conclusion}


The costly nature of DST data collection has been a limiting factor for the domain diversity of existing datasets for years. By introducing the first automatic data generation method capable of creating new domains and slot definitions for DST, this work both reveals and alleviates a performance bottleneck caused by the limited domain coverage of existing DST data. Training on the synthetic, domain-diverse \DOT dataset produces substantial performance gains (e.g. +$6.7\%$ average JGA) for zero-shot DST, and this performance gain is stable even when testing on domains with no similar analog in synthetic data. These results show the power of domain diversity for training zero-shot DST models, as it allows our models to achieve competitive or better performance to LLM-based DST approaches with over $13.5 \times$ the parameters.

The success of our data generation approach also demonstrates the potential of LLM-based data generation to alleviate the high costs of traditional data collection. Our work marks a pioneering step in the creation of similar fully automatic data generation approaches.  By continuing to improve the diversity and correctness of synthetic datasets, we anticipate even greater advancements in zero-shot DST performance, driving the development of more robust and adaptable dialogue systems. We look forward to future research and application development in task-oriented dialogue that builds upon our experimental insights and released models and data.

\section{Limitations}
\label{sec:limitations}

\paragraph{Redundancy of Slot Types} Although our presented data generation method successfully produces useful training data for zero-shot DST, it is important to note that this method does not produce a set of slot definitions where each slot is semantically unique. Our method attempts to maintain some consistency in tracking slots by modelling when requested slots are filled by a value. However, apart from tracking requested slots, slot-value update labels are generated relatively independently and without the notion of a centralized slot schema. This results in some cases, particularly across different dialogues belonging to the same domain, where slot labels are created with similar semantic meanings but different surface forms for slot names and descriptions. For training a zero-shot DST model this limitation is not an issue, since zero-shot DST models are expected to adapt to any provided slot name and definition to identify the correct value from the dialogue. However, the issue of inconsistent slot naming and lack of a centralized slot schema prevents datasets generated with our method from being used directly for few-shot training or DST evaluation. 

\paragraph{Noise in Silver State Labels} Since our data generation technique is fully automatic, it is expected that some noisy silver labels of dialogue state occur. The $82.0$\% slot-value correctness rate judged by our human annotators is interpretable as about 1 in 5 noisy slot-values. The limitation of this noise is that our experimental estimates of the impact of training data domain diversity on zero-shot DST are almost certainly under-estimates, as models trained on \DOT were trained to predict this noise. Ideally, a dataset of similar diversity to \DOT but with gold dialogue state labels would be used in our experiments; however, no such dataset exists, which is one of the primary motivations of our work. Our work thus serves as an investigation into the relationship between training domain diversity and zero-shot DST performance, but not one that conclusively quantifies this relationship. Future work should aim to reduce the noise in automatically generated DST labels or find more cost-efficient traditional data collection methods in order to achieve better experimental accuracy for measuring the impact of training domain diversity and in order to train higher-quality models.

\section{Ethical Considerations}
\label{sec:ethical_considerations}

 \paragraph{Risks} of this work are minimal; one risk introduced is through the use of GPT models to generate dialogue data, since it is theoretically possible for language model generations to populate synthetic dialogues with personal information of real people gathered from their training data. We believe the risk of this is low; after manually reviewing hundreds of dialogues in our \DOT data, we observe that most potentially sensitive information is generated by GPT in anonymized form (e.g. the phone number 555-5555).

 \paragraph{Languages} used in this work are restricted to English, since it was required for all the authors to understand model outputs during prompt development and error analysis. The methodology presented in this work fundamentally language-agnostic however, and can be adapted to new languages by translating prompts. Since \DOT is generated with a fully automatic method, analogous datasets in new languages can be created easily after prompt translation.

\bibliography{custom}

\clearpage
\newpage
\appendix

\section{Related Work in DST Data Generation}
\label{appx:related_work_for_dst_data_generation}

This section reviews previous work in DST data generation and augmentation, which targets few-shot DST. The theme of these works is to leverage a set of few shots as a seed set of examples used to generate additional synthetic examples in the target domain. By doing so, a limited set of training examples can be augmented for more robust DST training in the target domain.

\paragraph{Lexical Diversification} Some early approaches use paraphrasing techniques to improve lexical diversity on the turn-level. \citet{quan_effective_2019} experiment in this direction with a variety of methods such as back-translation and synonym replacement, and \citet{yin_dialog_2020} use a reinforcement learning approach to learn to replace token spans with paraphrases. These works demonstrate the potential of data augmentation to improve existing training resources, but their focus on paraphrasing fundamentally limits the extent to which the original data can be altered since the goal is to maintain the semantic content of original examples.

\paragraph{Semantic Diversification} Other approaches look to improve the generalizability of trained DST models to handle new values and dialogue contexts by modifying the semantic content of original dialogues. \citet{summerville_how_2020} focus specifically on the problem of DST models' ability to generalize to new slot values, using external corpora to augment training data with with additional values for open-ended slot types. \citet{lai_controllable_2022} synthesize new training examples by generating a new response to the context of existing dialogues. Their response generator is conditioned on the dialogue act and state, but is given a new dialogue act and state during augmentation to increase the semantic diversity of the training pool. These works successfully augment the lexical and semantic content of DST training data on the turn- or slot-value-level.

\paragraph{Dialogue Reconstruction} Some works augment existing data by synthesizing entirely new dialogues from an initial seed set. Three works explore methods that take advantage of the state representations in DST data to create a state transition graph, and then generate entirely new dialogues by traversing transition paths that are not represented in the initial dataset \cite{aksu_n-shot_2022, aksu_velocidapter_2021, campagna_zero-shot_2020}. Once a new state transition path for a synthetic dialogue is sampled from the transition graph, the turns from the original dialogues corresponding to each transition are used as templates and filled with new slot values to produce a final natural language dialogue. This approach introduces new variations in the structure and content of training data. However, the synthetic dialogues produced will share many of the same features as the original seed data, especially due to the reliance on templates. \citet{mehri_lad_2022} use a similar approach but eliminate the reliance on seed dialogues by using slot schema specification to create the state transition graph, and GPT-3 is used to paraphrase each template-generated turn to be more natural and coherent. It is difficult to evaluate the efficacy of their method however, since less-common evaluation data MixSNIPS/MixATIS \cite{qin_agif_2020} are used making comparison to related work difficult.

\paragraph{Full Dialogue Generation} Three recent works generate new DST data by training PLMs to generate new dialogues from a task goal and schema definition. \citet{kim_neuralwoz_2021} trained a dialogue generator model to produce dialogues given a goal, schema, and queryable database of schema values, and trained separate dialogue state labeler model to label the generated dialogues with dialogue states. \citet{mohapatra_simulated_2021} train a pipeline of separate PLMs to model a user response generator, user response selector, dialogue state generator, system response generator, and system responses selector. \citet{wan_unified_2022} similarly trained separate PLMs for to simulate user and system agents. They demonstrated improved transfer to generating synthetic data on low-resource target domains by pre-training their simulation agents on 12 different training data from previous work. All three of these approaches target low-resource DST by training their dialogue generation models on a limited amount of in-domain data, then train the DST model on synthetically generated data. Their results demonstrate the power of using PLMs to generate data to domains where substantial training resources are unavailable.

\section{Prompts}
\label{appx:prompts}

Eliciting high-quality generations from an LLM on a particular task requires finding a suitable prompt. The prompt is the token sequence input to the LLM that includes both task-specific instructions and a formatted linearization of all inputs needed to complete one task sample. Searching for a prompt that maximizes task performance can be done manually or using automatic or semi-automatic search methods \cite{prasad_grips_2023}. For complex tasks, multiple prompts can be used that decompose the task into more manageable subtasks. Due to the exploratory nature of our investigation into diverse DST data generation, we develop prompts through a manual development process where generations are hand-checked for quality. This allows us to quickly try different strategies for writing prompt instructions and breaking the data generation pipeline into subtasks. The prompts developed for the data generation pipeline (\textsection\ref{sec:generating_dst_data}) are shown in Figures \ref{fig:prompt-domains} - \ref{fig:prompt-slot-description}.

\section{Domains}
\label{appx:domains}

To show the kinds of scenario descriptions generated for \DOT (\textsection\ref{ssec:scenario-derivation}) that were used as task domains, we randomly sample 40 scenario descriptions from the complete set of 1,003 and present them in Table \ref{tab:domains-by-split}.

\section{Implementation Details}
\label{appx:implementation-details}

\paragraph{Llama-13B-Chat} is a 13 billion parameter decoder-only transformer model trained on a variety of long-form texts, then further trained on instruction data using the Reinforcement Learning from Human Feedback (RLHF) technique \cite{ouyang_training_2022}. Due to the computational expense of its 13B parameter size, the model was quantized using QLoRA \cite{dettmers_qlora_2023}, which uses 4-bit \texttt{nf4} quantization, and freezes the base model parameters while only training the parameters of a Low-Rank Adapter (LoRA) \cite{hu_lora_2022} of rank 32. Training used a learning rate of $2e-5$, and a batch size of $256$, with no dropout or weight decay.

\paragraph{T5-11B} \cite{raffel_exploring_2020} is a 11 billion parameter encoder-decoder transformer model trained on a variety of sequence-to-sequence tasks such as summarization and translation. The T5 1.1 variant was used, following \citet{gupta_show_2022}. QLoRA training used a rank of 32, alpha of 64, with a learning rate of $1e-2$ and batch size of $256$, with no dropout or weight decay. Full fine-tuning used a learning rate of $1e-3$ with weight decay $5e-3$.

\section{Error Analysis}
\label{appx:error-analysis}

\begin{table}[htb!]
    \small
    \centering
    \resizebox{\columnwidth}{!}{%
    \begin{tabular}{p{3em}|p{12em}|cc}
    \toprule
        \multicolumn{1}{c|}{\multirow{2}{*}{\bf Error}} & \multicolumn{1}{c|}{\multirow{2}{*}{\bf Definition}}  & \bf Stage & \bf Stage \\
        & & \bf 1 & \bf 2 \\
        \midrule
        Agent Value Miss & No value is outputted for the indicated slot, even though the information is present in the system's turns. & 13 & 13 \\
        
        \midrule
        No Preference & Indications of no preference are inappropriately understood, either by failing to recognize when no preference is given or by incorrectly interpreting an indication of no preference from the dialogue. & 13 & 9 \\
        
        \midrule
        Value Change & The appropriate value for the indicated slot has been updated in the dialogue turn, but the predicted value remains as the original. & 10 & 19 \\
        
        \midrule
        Halluc- ination & A value is predicted for the indicated slot that does not exist in the dialogue. & 9 & 5 \\
        
        \midrule
        Miss & No value is outputted for the indicated slot, even though the information is present in the user's turns. & 7 & 13 \\
        
        \midrule
        Wrong Value & Information in the dialogue is incorrectly attributed to the indicated slot. & 6 & 8 \\
        
        \midrule
        Other & Errors not explained by any of the other error patterns. & 16 & 11 \\
        
        \midrule
        Correct & The predicted value for the indicated slot is correct, but is missing from the gold annotations in MultiWOZ due to an annotation mistake. & 26 & 22 \\
 
    \bottomrule
    \end{tabular}%
    }
    \caption{Error analysis on 100 randomly sampled erroneous outputs on MultiWOZ 2.4 of the best-performing finetuned Llama-13B-Chat model with QLoRA training (Stage 2) and the same model trained only on \DOT (Stage 1), before fine-tuning on MultiWOZ.}
    \label{tab:dsg5k-pretrained-0shot-error-analysis}
    \vspace{-3ex}
\end{table}

The impact of diverse DST training data is further investigated by conducting an error analysis on 100 randomly sampled errors from the best-performing Llama2 +\texttt{\DOT} +\texttt{ICL} model. The model was evaluated for both Stage 1 (\DOT training only) and Stage 2 (subsequent training on MultiWOZ), and the results of the error analysis can be seen in Table \ref{tab:dsg5k-pretrained-0shot-error-analysis}. As expected, some of the errors made by these models are due to slot semantics specific to the MultiWOZ task that are difficult to encode in a single-sentence slot description. For example, the \texttt{dontcare} value (represented as \texttt{any} to the model) is a frequent source of errors, as the model consistently overpredicts it in the Hotel domain. Many errors also stem from a slot being filled with a wrong value that does indeed appear in the dialogue, but does not quite fit the specifics of the definition of the MultiWOZ slot. However, the majority of errors made are due to limitations in the training formulation using the synthetic dataset. For example, the dialogues generated by GPT-3.5 rarely include corrections or clarifications where slot value would change, resulting in consistent errors when the user speaker changes their mind or self-corrects in MultiWOZ. Also, the military time format used in MultiWOZ for time slots was a consistent source of hallucinations, as this format rarely or never appears in the synthetic \DOT data. Finally, the models frequently missed slot values entirely, particularly when the value originated from the system travel agent speaker.

\begin{figure*}[t]
   \centering
   \small \begin{verbatim}
    List 100 diverse examples of everyday tasks that require talking to another person. 
    Format each list item like:

    N. <Role of person 1> talks to <role of person 2> in order to <task goal>\end{verbatim}
   \caption{GPT-3.5 prompt for generating dialogue scenarios/domains.}
   \label{fig:prompt-domains}
\end{figure*}

\begin{figure*}[t]
   \centering
   \small \begin{verbatim}
    List examples of as many different types of information as you can that would be
    shared during the dialogue scenario: {domain}\end{verbatim}
   \caption{GPT-3.5 prompt for generating a list of information types for each dialogue domain.}
   \label{fig:prompt-domain-info-types}
\end{figure*}

\begin{figure*}[t]
   \centering
   \small \begin{verbatim}
    Dialogue Scenario:
    {domain}

    Information Types:
    {info types}

    Write a dialogue for the above Dialogue Scenario. Include specific examples of the 
    Information Types above being shared and implied throughout the conversation. 
    Make up actual names/values when specific information examples are shared.\end{verbatim}
   \caption{GPT-3.5 prompt for generating a dialogue for a given task domain.}
   \label{fig:prompt-domain-dialogue}
\end{figure*}

\begin{figure*}[t]
   \centering
   \small \begin{verbatim}
    Two people, {speaker} and {listener}, are having a dialogue in which the 
    following was just said:

    {dialogue context}
    {speaker}: {last turn}

    Please break down and summarize all the information in what {speaker} just
    said into as many question-answer pairs as you can. Each question-answer pair
    should be short, specific, and focus on only one piece of information or value.

    For information {speaker} shared, use the question-answer pair format:

    {listener}: <question>
    {speaker}: <answer>

    For information {speaker} requested or indicated not knowing,
    use the answer "Unknown." in a question-answer pair format like:

    {speaker}: <question>
    {listener}: Unknown.


    {answered qa pairs}\end{verbatim}
   \caption{GPT-4 prompt for generating question-answer pairs for a dialogue context.}
   \label{fig:prompt-qa-pairs}
\end{figure*}

\begin{figure*}[t]
   \centering
   \small \begin{verbatim}
    Two people, {speaker} and {listener}, are having a dialogue in which the
    following was just said:

    {dialogue context}
    {speaker}: {last turn}

    Please identify the information or values {speaker} gave as short answers to the
    following questions (use the answer "Unknown." if the question is not answered by 
    {speaker} in the dialogue):

    {unanswered qa questions}\end{verbatim}
   \caption{GPT-4 prompt for answering questions from the previous turn that were not previously answered.}
   \label{fig:prompt-qa-answers}
\end{figure*}

\begin{figure*}[t]
   \centering
   \small \begin{verbatim}
    {qa pairs}

    Translate each question above into variable names. 
    Each label should be very short, usually one or two words, 
    but specific to the details of the question. Write each question before
    translating it into a variable name, in the format:

    <question> -> <variable name>\end{verbatim}
   \caption{GPT-3.5 prompt for translating questions into slot names.}
   \label{fig:prompt-slotname-translation}
\end{figure*}

\begin{figure*}[t]
   \centering
   \small \begin{verbatim}
    {qav tuples}

    Translate each answer to the above questions into a value for the
    corresponding variable. Values should be short, usually one word, 
    very short phrase, number, span, category, score, boolean, list, 
    or other value. Copy each answer before translating it into a value,
    in the format:

    Question: <question>
    Variable: <variable>
    Answer: <answer>
    Value: <value>\end{verbatim}
   \caption{GPT-3.5 prompt for translating answers into slot values.}
   \label{fig:prompt-value-translation}
\end{figure*}

\begin{figure*}[t]
   \centering
   \small \begin{verbatim}
    {slots with corresponding questions and values}

    For each Info Type above, write a comma-separated list of all Possible Values 
    (if there are many Possible Values, write ", etc." after a few examples), 
    and a short phrase as a description for each Info Type. Use the format:

    Info Type: <info type>
    Possible Values: <value 1>, <value 2>, <value 3>
    Description: <phrase>\end{verbatim}
   \caption{GPT-3.5 prompt for generating descriptions and value examples for each slot.}
   \label{fig:prompt-slot-description}
\end{figure*}

\begin{table*}[htbp!]
    \centering
    \resizebox{\textwidth}{!}{%
    \begin{tabular}{l}
    \toprule
Parent talks to pediatrician in order to schedule vaccinations. \\
Pet owner talks to veterinarian in order to schedule a check-up \\
Event organizer talks to security personnel in order to ensure safety at an event \\
Presenter talks to audio technician in order to test the sound system before a conference \\
Bartender talks to bouncer in order to assist with maintaining safety and order in a bar or club \\
Performer talks to stage crew in order to coordinate a show \\
Retail sales associate talks to customer in order to assist with an item purchase \\
Executive talks to assistant in order to delegate tasks and schedule appointments. \\
Hair stylist talks to bride in order to plan a wedding up-do \\
Parent talks to teacher about afterschool programs. \\
Parent talks to nutritionist in order to receive guidance on healthy eating for their family \\
Blogger talks to other bloggers in order to collaborate on blog content. \\
Coworker talks to mentor in order to receive guidance on career development. \\
Homeowner talks to landscaper in order to plant new flowers. \\
Mover talks to customer in order to move their belongings \\
Fortune teller talks to client in order to provide a fortune prediction. \\
Proofreader talks to author in order to check for grammatical errors and typos in writing \\
Coworker talks to coworker in order to discuss a workplace policy. \\
Magazine editor talks to writer in order to edit their piece. \\
Talent agent talks to actor in order to develop a career plan. \\
Comedian talks to event planner in order to discuss comedy act material \\
Participant talks to moderator in order to ask a question during a session. \\
Significant other talks to partner in order to make plans for the future. \\
Passenger talks to flight attendant in order to ask for an extra pillow. \\
Survivor talks to counselor in order to receive support after traumatic event. \\
Animal behaviorist talks to zookeeper in order to observe and analyze animal behavior patterns \\
Freelance writer talks to editor in order to pitch article ideas \\
Tourist talks to tour guide in order to learn about a city's history. \\
Manager talks to HR representative in order to review job applications \\
Job seeker talks to employment agency in order to find a job. \\
Legal assistant talks to client in order to assist with legal paperwork \\
Pets blogger talks to subscribers in order to provide information about pets \\
Salesperson talks to manager in order to receive training \\
Motivational speaker talks to audience in order to inspire them \\
Dentist talks to insurance adjuster in order to find out what procedures are covered \\
Box office attendant talks to patron in order to sell tickets. \\
Boss talks to employee in order to give feedback on a project. \\
Attendee talks to speaker in order to say thank you after a presentation. \\
Project manager talks to stakeholders in order to provide updates \\
Postman talks to colleague to coordinate deliveries \\

\bottomrule
    \end{tabular}
    }
    \caption{Random sample of 40 scenario descriptions generated for \DOT (\textsection\ref{ssec:scenario-derivation}) to serve as task domains.}
    \label{tab:domains-by-split}
\end{table*}

\end{document}